\documentclass[runningheads]{llncs}
\usepackage{graphicx}
\usepackage{amsmath,amssymb} % define this before the line numbering.
\usepackage{url}
\usepackage{cite}
\usepackage{color}
\usepackage{comment}
\usepackage{epsfig}
\usepackage{graphicx}
\usepackage{amsmath}
\usepackage{amssymb}
\usepackage{multirow}
\usepackage{booktabs}
\usepackage{algorithm}
\usepackage{algorithmic}
\usepackage{arydshln}
\usepackage{threeparttable}
\usepackage{colortbl}
\usepackage{enumitem}
\usepackage{siunitx}
\usepackage{bm}
\usepackage{enumitem}
\usepackage{bm}
\usepackage{tablefootnote}
\usepackage{array}
\usepackage{colortbl}
\usepackage{caption}
\usepackage{dblfloatfix}
\usepackage{subcaption}
\usepackage{wrapfig,lipsum,booktabs}
\usepackage{marvosym}
\usepackage{amssymb}% http://ctan.org/pkg/amssymb
\usepackage{pifont}% http://ctan.org/pkg/pifont

\newcommand{\cmark}{\ding{51}}%
\newcommand{\xmark}{\ding{55}}%
\definecolor{mygray}{gray}{.9}
\definecolor{ggray}{RGB}{127,127,127}
\definecolor{reda}{RGB}{192,0,0}
\definecolor{redb}{RGB}{217,148,143}
\definecolor{myyellow}{RGB}{190,144,0}
\definecolor{mygreen}{RGB}{80,100,40}
\definecolor{myblue}{RGB}{30,90,100}
\newcommand{\etal}{\textit{et al}.}
\newcommand{\ie}{\textit{i}.\textit{e}.}
\newcommand{\eg}{\textit{e}.\textit{g}.}
\newcommand{\etc}{\textit{etc}.}
\newcommand{\wrt}{\textit{w.r.t}.}

\makeatletter
\newcommand{\thickhline}{%
    \noalign {\ifnum 0=`}\fi \hrule height 1pt
    \futurelet \reserved@a \@xhline
}
\usepackage[pagebackref=true,breaklinks=true,letterpaper=true,colorlinks,bookmarks=false]{hyperref}
\usepackage[width=122mm,left=12mm,paperwidth=146mm,height=193mm,top=12mm,paperheight=217mm]{geometry}
\begin{document}
% \linenumbers
\pagestyle{headings}
\mainmatter
\def\ECCVSubNumber{3387}  % Insert your submission number here
%Deep Semantic Relation Mining for
%\title{Weakly$_{\!}$ Supervised$_{\!}$ Semantic$_{\!}$ Segmentation$_{\!}$ with$_{\!}$ Cross-Image$_{\!}$ Semantic$_{\!}$ Relations$_{\!}$} % Replace with your title
\title{Mining Cross-Image Semantics for Weakly Supervised Semantic Segmentation}

\titlerunning{Mining Cross-Image Semantics for WSSS}

\authorrunning{G. Sun, W. Wang, J. Dai, L. Van Gool}

\author{Guolei Sun$^{1}$\and
\Letter Wenguan Wang$^{1}$\and
Jifeng Dai$^{2}$\and
Luc Van Gool$^{1}$}
\institute{\small $^{1}$ ETH Zurich~~$^{2}$ SenseTime Research\\$^{3}$ Qing Yuan Research Institute, Shanghai Jiao Tong University\\
\url{https://github.com/GuoleiSun/MCIS_wsss}
}

\maketitle
% \vspace{-10pt}
\begin{abstract}
%Current top-leading weakly supervised semantic segmentation (WSSS) solutions rely on the attention maps from image classifier.
%Object attention maps generated by image classifiers are usually used as priors for weakly-supervised segmentation approaches
%Recent state-of-the-art methods on this problem first infer the sparse and discrimi- native regions for each object class using a deep classifica- tion network, then train semantic a segmentation network using the discriminative regions as supervision.
This paper studies the problem of learning semantic segmentation from image-level supervision only. Current popular solutions leverage object localization maps from classifiers as supervision signals, and struggle to make the localization maps capture more complete object content. Rather than previous efforts that primarily focus on \textit{intra-image} information, we address the value of \textit{cross-image} semantic relations for comprehensive object pattern mining. To achieve this, two neural co-attentions are incorporated into the classifier to complimentarily capture cross-image semantic similarities and differences. In particular, given a pair of training images, one co-attention enforces the classifier to recognize the common semantics from co-attentive objects, while the other one, called contrastive co-attention, drives the classifier to identify the unshared semantics from the rest, uncommon objects. This helps the classifier discover more object patterns and better ground semantics in image regions. In addition to boosting object pattern learning, the co-attention can leverage context from other related images to improve localization map inference, hence eventually benefiting semantic segmentation learning. More essentially, our algorithm provides a unified framework that handles well different WSSS settings, \ie, learning WSSS with (1) precise image-level supervision only, (2) extra simple single-label data, and (3) extra noisy web data. It sets new state-of-the-arts on all these settings, demonstrating well its efficacy and generalizability. Moreover, our approach ranked $1^{st}$ place in the Weakly-Supervised Semantic Segmentation Track of CVPR2020 Learning from Imperfect Data Challenge.

%, Co-Attention, Semantic Relation
\end{abstract}

\let\thefootnote\relax\footnotetext{\Letter~Corresponding author: Wenguan Wang (wenguanwang.ai@gmail.com).}
%\vspace{-25pt}
\section{Introduction}
%\vspace{-5pt}
%With the widespread adoption of deep neural networks in computer vision, remarkable progress has been made in the field. For example,
Recently, modern deep learning based semantic segmentation models\!~\cite{chen2017deeplab,deeplabv3plus2018}, trained with massive manually labeled data, achieve far better performance than before.
However, the fully supervised learning paradigm has the main limitation of requiring intensive manual labeling effort, which is particularly expensive for annotating pixel-wise ground-truth for semantic segmentation. %Due to that, there is lately
Numerous efforts are motivated to develop semantic segmentation with weaker forms of supervision, such as bounding boxes\!~\cite{papandreou2015weakly}, scribbles\!~\cite{lin2016scribblesup},  points\!~\cite{bearman2016s}, and image-level labels\!~\cite{pathak2014fully}, \etc ~Among them, a prominent and appealing trend is using only image-level labels to achieve weakly supervised semantic segmentation (WSSS), which demands the least annotation efforts and is followed in this work.

To tackle the task of WSSS with only image-level labels, current popular methods are based on network visualization techniques\!~\cite{zeiler2014visualizing,cam2016learning}, which discover discriminative regions that are activated for classification. These methods use image-level labels to train a classifier network, from which class-activation maps are derived as pseudo ground-truths for further supervising pixel-level semantics learning. However, it is commonly evidenced that the trained classifier tends to over-address the most discriminative parts rather than entire objects, which becomes the focus of this area. Diverse solutions are explored, typically adopting: \textit{image-level} operations, such as region hiding and erasing\!~\cite{kumar2017hide,wei2017object}, \textit{regions growing} strategies that expand the initial activated regions\!~\cite{sec2016,wang2018weakly}, and \textit{feature-level} enhancements that collect multi-scale context from deep features\!~\cite{wei2018revisiting,lee2018robust}.

\begin{figure}[t]
%%tr = 0.006, ts = 0.008
  \centering
      \includegraphics[width=0.99\linewidth]{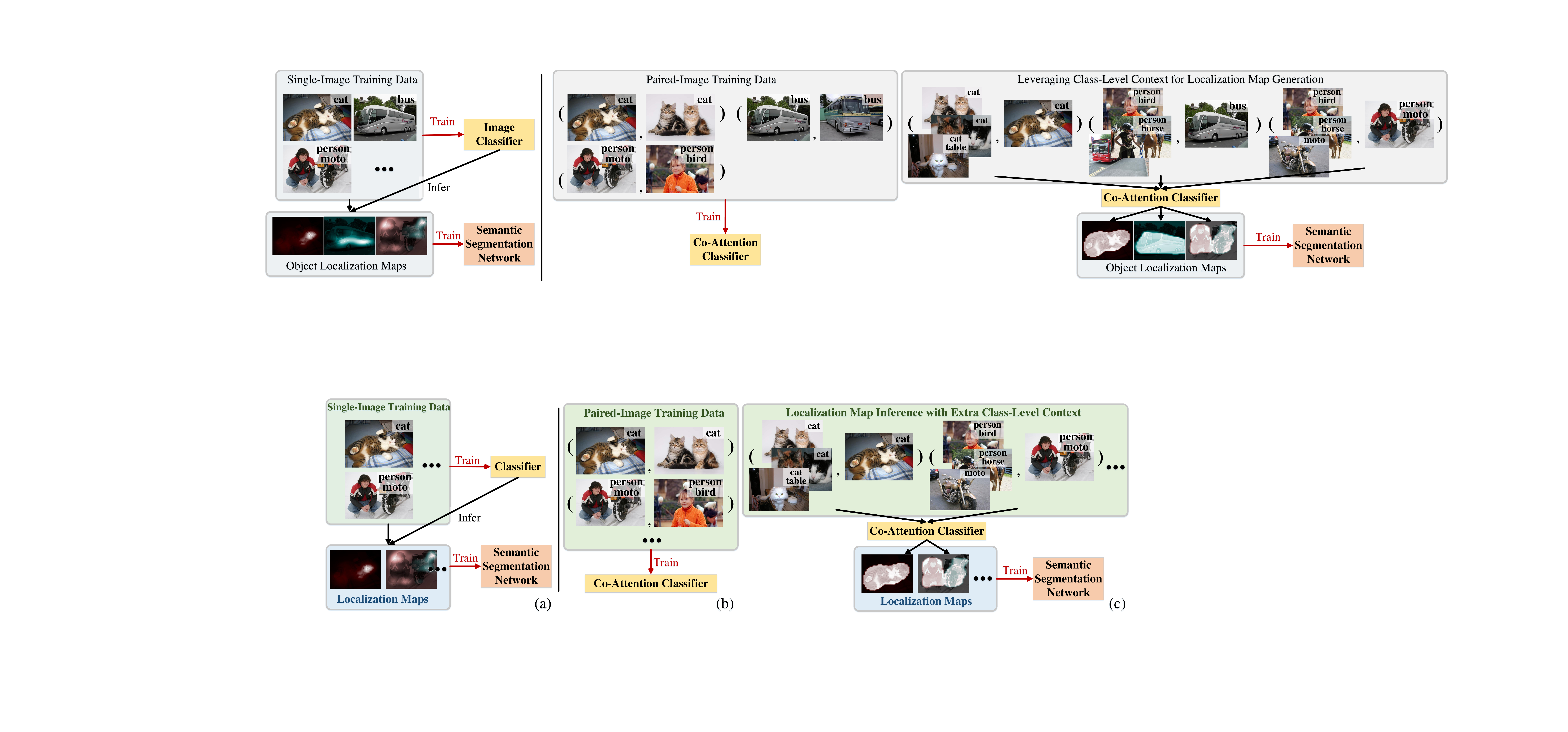}
      %\put(-89,77){\fontsize{6pt}{6.2pt}\selectfont (Eq.~\!\ref{eq:Pseudo})}
\vspace{-3pt}
\captionsetup{font=small}
\caption{\small{(a) Current WSSS methods only use single-image information for object pattern discovering. (b-c) Our co-attention classifier leverages cross-image semantics as class-level context to benefit object pattern learning and localization map inference.
}
}
\label{fig:overview}
\vspace{-12pt}
\end{figure}

These efforts generally achieve promising results, which demonstrates the importance of discriminative object pattern mining for WSSS. However, as shown in Fig.~\ref{fig:overview}(a), they typically use only single-image information for object pattern discovering, ignoring the rich semantic context among the weakly annotated data. For example, with the image-level labels, not only the semantics of each individual image can be identified, the cross-image semantic relations, \ie, two images whether sharing certain semantics, are also given and should be used as cues for object pattern mining. Inspired by this, rather than relying on \textit{intra-image} information only,  we further address the value of \textit{cross-image} semantic correlations for  complete object pattern learning and effective class-activation map inference (see Fig.~\ref{fig:overview}(b-c)). In particular, our classifier is equipped with a differentiable co-attention mechanism that addresses semantic  homogeneity
and difference understanding across training \textit{image pairs}.  More specifically, two kinds of co-attentions are learned in the classifier. The former one aims to capture cross-image common semantics, which enables the classifier to better ground the common semantic labels over the co-attentive regions. The latter one, called contrastive co-attention, focuses on the rest, unshared semantics, which helps the classifier better separate semantic patterns of different objects. These two co-attentions work in a cooperative and complimentary manner, together making the classifier understand object patterns more comprehensively.

In addition to benefiting object pattern learning, our co-attention provides an efficient tool for precise localization map inference  (see Fig.~\ref{fig:overview}(c)). Given a training image, a set of related images (\ie, sharing certain common semantics) are utilized by the co-attention for  capture richer context and generate more accurate localization maps.  Another advantage is that our co-attention based classifier learning paradigm brings an efficient data augmentation strategy, due to the use of training image pairs. Overall,  our co-attention boosts object discovering during both the classifier's training phase as well as localization map inference stage. This provides the possibility of obtaining more accurate pseudo pixel-level annotations, which facilitate final semantic segmentation learning.

Our algorithm is a unified and elegant framework, which generalizes well different WSSS settings. Recently, to overcome the inherent limitation in WSSS without additional human supervision, some efforts resort to extra image-level supervision from simple single-class data readily available from other existing datasets~\cite{pinheiro2015image,li2019attention}, or cheap web-crawled data~\cite{shen2017weakly,hong2017weakly,shen2018bootstrapping,wei2016stc}. Although they improve the performance to some extent, complicated techniques, such as energy function optimization~\cite{hong2017weakly,tokmakov2016weakly}, heuristic constraints~\cite{shen2018bootstrapping}, and curriculum learning~\cite{wei2016stc}, are needed to handle the challenges of domain gap and data noise, restricting their utility.  However, due to the use of paired image data for classifier training and object map inference, our method has good tolerance to noise. In addition, our method also handles domain gap naturally, as the co-attention effectively addresses domain-shared object pattern learning and achieves domain adaption as a part of co-attention parameter learning. We conduct extensive experiments on PASCAL VOC 2012~\cite{everingham2015pascal}, under three WSSS settings, \ie, learning WSSS with \textbf{(1)} PASCAL VOC image-level supervision only, \textbf{(2)} extra simple single-label data, and \textbf{(3)} extra web data. Our algorithm sets state-of-the-art on each case, verifying its effectiveness and generalizability. Our method also ranked 1$^{st}$ place in the Weakly-supervised Semantic Segmentation Track of CVPR2020 Learning from Imperfect Data (LID) Challenge\!~\cite{lid2020} (LID$_{20\!}$), outperforming other competitors by large margins.

Our contributions are three-fold. \textbf{(1)} We address the value of cross-image semantic correlations for complete object pattern learning as well as object location inference, which is achieved by a co-attention classifier that works over paired training samples. \textbf{(2)} Our co-attention classifier mines semantic cues in a more comprehensive manner. In addition to single-image semantics, it mines complimentary supervision from cross-image semantic similarities and differences by the co-attention and contrastive co-attention, respectively. \textbf{(3)} Our approach is general enough to learn WSSS with precise image-level supervision, or with extra simple single-label, or even noisy web-crawled data. It solves inherent challenges of different WSSS settings elegantly, and shows promising results consistently.

%for the first time,
%This leads to the potential of boosting
%generalization ability of a classification network in both
%domain adaptation and domain generalization settings.

% label \cite{lee2019frame}

%Class-level, image-wise

%In addition, the variations introduced by the image transformation operations allow a classifier to
 %activate different parts of objects, which provides the possibility
%of obtaining better pixel-level annotations.
%
%We propose

%However, different to previous efforts that rarely consider the rich semantic correlations among training samples, we

\vspace{-5pt}
\section{Related Work}\label{sec:rw}
\vspace{-3pt}
% In this section, we first briefly review recent advances in weakly supervised semantic segmentation and then introduce differentiable neural attention mechanisms.
%In this section, we briefly review recent advances in weakly supervised semantic segmentation (WSSS), followed by differentiable neural attention mechanisms.

\noindent\textbf{Weakly Supervised Semantic Segmentation.} Recently, lots of WSSS methods have been proposed to alleviate labeling cost. Various weak supervision forms have been explored, such
as bounding boxes \cite{papandreou2015weakly, dai2015boxsup}, scribbles~\cite{lin2016scribblesup}, point supervision~\cite{bearman2016s}, \etc~ Among them, image-level supervision, due to its less annotation demand, gains most attention and is also adopted in our approach.

Current popular solutions for WSSS with image-level supervision rely on network visualization techniques~\cite{zeiler2014visualizing,cam2016learning}, especially the Class Activation Map (CAM)~\cite{cam2016learning}, which discovers image pixels that are informative for classification. However, CAM typically only identifies small discriminative parts of objects, making it not an ideal proxy ground-truth for  semantic segmentation training. Therefore, numerous efforts are made towards expanding the CAM-highlighted regions to the whole objects. In particular, some representative approaches make use of \textit{image-level} hiding and erasing operations to drive a classifier to focus on different parts of objects~\cite{li2018tell,kumar2017hide,wei2017object}.
A few ones instead resort to a \textit{regions growing} strategy, \ie, view the CAM-activated regions as initial ``seeds'' and gradually grow the seed regions until cover the complete objects~\cite{sec2016,wang2018weakly,dsrg2018,psa2018}. Meanwhile, some researchers
investigate to directly enhance the activated regions on \textit{feature-level}~\cite{wei2018revisiting,lee2018robust,lee2019ficklenet}. When constructing CAMs, they collect multi-scale context, which is achieved by dilated convolution~\cite{wei2018revisiting}, multi-layer feature fusion~\cite{lee2018robust}, saliency-guided iterative training~\cite{wang2018weakly}, or stochastic feature selection~\cite{lee2019ficklenet}. Some others accumulate CAMs from multiple training phases~\cite{oaa2019}, or self-train a difference detection network to complete the CAMs with trustable information~\cite{ssdd2019}. In addition, a recent trend is to utilize class-agnostic saliency cues to filter out background responses~\cite{wei2017object,dsrg2018,wang2018weakly,li2018tell,wei2018revisiting,lee2019ficklenet,fan2020cian} during localization map inference.\!\!

%,
Since the supervision provided in above problem setting is so weak, another category of approaches explores to leverage more image-level supervision from other sources. There are mainly two types: (1) exploring simple and single-label examples~\cite{pinheiro2015image,li2019attention} (\eg, images from existing datasets \cite{russakovsky2015imagenet,griffin2007caltech}); or (2) utilizing near-infinite yet noisy web-sourced image~\cite{shen2017weakly,hong2017weakly,shen2018bootstrapping,wei2016stc} or video~\cite{tokmakov2016weakly,hong2017weakly,lee2019frame} data (also referred as \textit{webly supervised semantic segmentation} \cite{jin2017webly}). In addition to the common challenge of domain gap between the extra data and target semantic segmentation dataset, the second-type methods need to handle data noise.

Past efforts only consider each image individually, while only few exceptions~\cite{shen2017weakly,fan2020cian} address cross-image information. \cite{shen2017weakly} simply applies off-the-shelf co-segmentation~\cite{joulin2010discriminative} over the web images to generate foreground priors, instead of ours encoding the semantic relations into network learning and inference. For \cite{fan2020cian}, although also exploiting correlations within image pairs, the core idea is to use extra information from a support image to supplement current visual representations. Thus the two images are expected to better contain the same semantics, and unmatched semantics would bring negative influences. In contrast, we view both semantic homogeneity and difference as informative cues, driving our classifier to more explicitly identify the common as well as unshared objects, respectively.  Moreover, \cite{fan2020cian} only utilizes single image to infer the activated objects, but our method comprehensively leverages the cross-image semantics in both classifier training and localization map inference stages. More essentially, our framework is neat and flexible, which is not only able to learn WSSS from clean image-level supervision, but general enough to naturally make use of extra noisy web-crawled or simple single-label data, contrarily to previous efforts which are limited to specific training settings and largely dependent on complicated optimization methods~\cite{hong2017weakly,tokmakov2016weakly} or heuristic
constraints~\cite{shen2018bootstrapping}.

\noindent\textbf{Deterministic Neural Attention.} Differentiable attention mechanisms enable a neural network to focus more on relevant elements of the input than on irrelevant parts.  With their popularity in the field of natural language processing \cite{vaswani2017attention,luong2015effective,cheng2016long,lin2017structured,paulus2017deep}, attention modeling  is rapidly adopted in various computer vision tasks, such as image recognition \cite{fu2019dual,mamc2018,wang2018non,hu2018squeeze,woo2018cbam}, domain adaptation \cite{wang2019transferable,zhang2019sequence}, human pose estimation \cite{wang2020hierarchical,chu2017multi,ye2016spatial}, reasoning among objects \cite{zhou2020cascaded,santoro2017simple}, and image generation \cite{zhu2019progressive,zhang2018self,xu2018attngan}. Further, co-attention mechanisms become an essential tool in many vision-language applications and sequential modeling tasks, such as visual question answering \cite{lu2016hierarchical,xiong2016dynamic,nguyen2018improved,yu2019deep}, visual dialog \cite{zheng2019reasoning,wu2018you}, vision-language navigation \cite{wang2019reinforced}, and video segmentation \cite{wang2019zero,lu2019see}, showing its effectiveness in capturing the underlying relations between different entities. Inspired by the general idea of attention mechanisms, this work leverages co-attention to mine semantic relations within training image pairs, which helps the classifier network learn complete object patterns and generate precise object localization maps. %classification tags on spatial visual domain. %Note that XX suggests a ``contrastive attention'' based network visualization strategy. It highlights a part of network units by suppressing responses from the others, which is totally different to ours.

% \vspace{-5pt}
% \noindent\textbf{Semantic Segmentation:}

\vspace{-7pt}
\section{Methodology}
\label{sec:method}
\vspace{-3pt}
%Fig. illustrates an overview of our WSSS model.
%Most previous works \cite{oaa2019,lee2019frame,psa2018,lee2019ficklenet} have the following pipeline: train a classification network, generate localization maps and pseudo ground-truth masks, and finally train fully supervised segmentation network. Our proposed method follows this standard pipeline. In the following, we first detaily describe the proposed co-attention classification network in \ref{co-cls}, and then explain how to train the network and generate localization maps in \ref{inference}. Finally, in \ref{full-seg} we briefly introduce the training of segmentatin network.
\noindent\textbf{Problem Setup.} Here we follow current popular WSSS pipelines: given a set of training images with image-level labels, a \textit{classification network} is first trained to discover corresponding discriminative object regions. The resulting \textit{object localization maps} over the training samples are refined as pseudo ground-truth masks to further supervise the learning of a \textit{semantic segmentation network}.

\noindent\textbf{Our Idea.} Unlike most previous efforts that treat each training image \textit{individually}, we explore cross-image
semantic relations as class-level context for understanding object patterns more \textit{comprehensively}. To achieve this, two neural co-attentions are designed. The first one drives the classifier to learn common semantics from the co-attentive object regions, while the other one enforces the classifier to focus on the rest objects for unshared  semantics classification. % When the paired training samples are two different images, the classifier is inspired to comprehensive semantic understanding. When the paired training samples are from a same image, transformations are pre-applied to make more robust object pattern modeling.
%In essence, our algorithm explicitly leverages \textit{class-level} context to learn and infer object patterns, instead of prior art only relying on \textit{intra-image} cues.

%Like most previous work
\vspace{-7pt}
\subsection{Co-attention Classification Network}\label{co-cls}
\begin{figure}[t]
%%tr = 0.006, ts = 0.008
  \centering
      \includegraphics[width=0.99\linewidth]{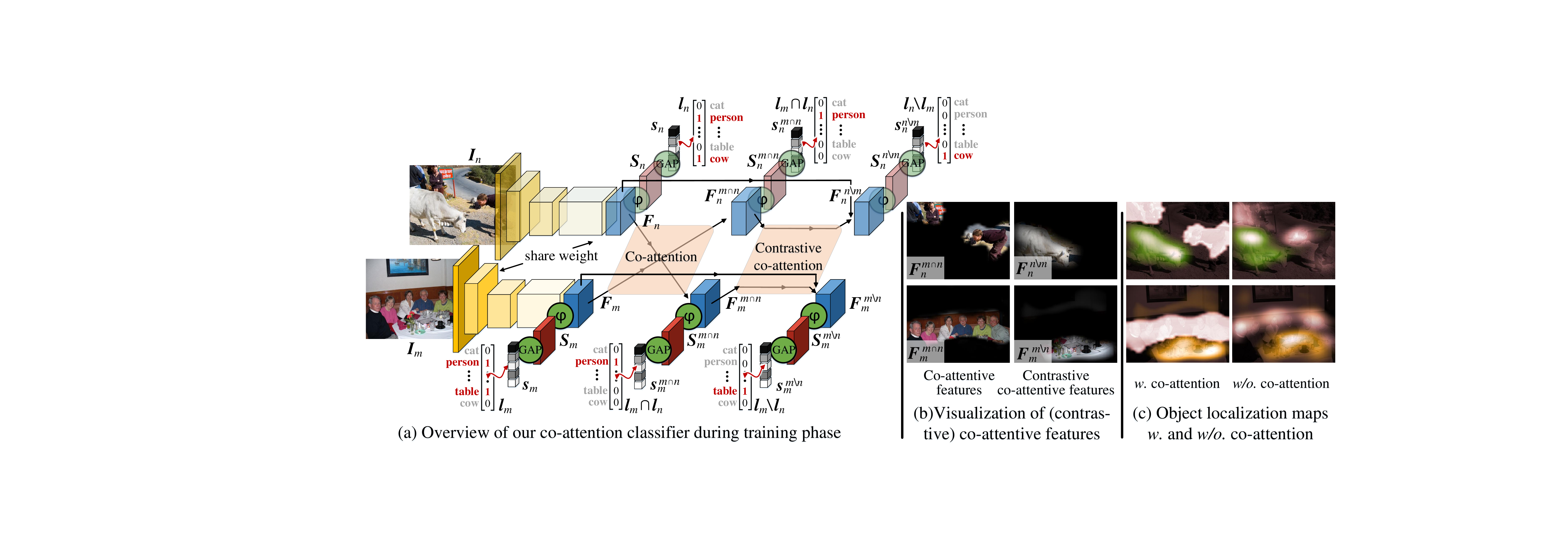}
      %\put(-89,77){\fontsize{6pt}{6.2pt}\selectfont (Eq.~\!\ref{eq:Pseudo})}
\vspace{-2pt}
\captionsetup{font=small}
\caption{\small{\textbf{(a)} In addition to mining object semantics from single-image labels, semantic similarities and differences between paired training images are both leveraged for supervising object pattern learning. \textbf{(b)} Co-attentive and contrastive co-attentive features complimentarily capture the shared and unshared objects. \textbf{(c)} Our co-attention classifier is able to learn object patterns more comprehensively. \textit{Zoom-in for details}.
}
}
\label{fig:framework}
\vspace{-8pt}
\end{figure}

\vspace{-3pt}
Let us denote the training data as $\mathcal{I}\!=\!\{(\bm{I}_n,\bm{l}_n)\}_n$, where $\bm{I}_n$ is the $n^{th}$ training image, and $\bm{l}_{n\!}\!\in\!\!\{0,1\}^K$ is the associated \textit{ground-truth} image label for $K$ semantic categories. As shown in Fig.~\ref{fig:framework}(a), image pairs, \ie, $(\bm{I}_m,\bm{I}_n)$, are sampled from $\mathcal{I}$ for training the classifier. After feeding $\bm{I}_m$ and $\bm{I}_n$ into the convolutional embedding part of the classifier, corresponding feature maps,  $\bm{F}_{m\!}\!\in~\!\!\mathbb{R}^{C\!\times\! H\!\times\!W\!}$ and $\bm{F}_{n\!}\!\in~\!\!\mathbb{R}^{C\!\times\! H\!\times\!W\!}$, are obtained, each with $H\!\times\!W$ spatial dimension and $C$ channels.

As in \cite{oaa2019,lee2019ficklenet,lee2019frame}, we can first separately pass $\bm{F}_{m}$ and $\bm{F}_{n}$ to a \textit{class-aware fully convolutional layer} $\varphi(\cdot)$ to generate \textit{class-aware activation maps}, \ie, $\bm{S}_{m\!}\!=\!\varphi(\bm{F}_{m})\!\in\!\mathbb{R}^{K\!\times\!H\!\times\!W\!}$ and $\bm{S}_{n\!}\!=\!\varphi(\bm{F}_{n})\!\in\!\mathbb{R}^{K\!\times\!H\!\times\!W\!}$, respectively. Then, we apply  \textit{global average pooling} (GAP) over $\bm{S}_{m}$ and $\bm{S}_{n}$ to obtain class score vectors $\bm{s}_{m\!}\!\in\!\mathbb{R}^{K\!}$ and $\bm{s}_{n\!}\!\in\!\mathbb{R}^{K\!}$ for $\bm{I}_{m\!}$ and $\bm{I}_n$, respectively. Finally, the \textit{sigmoid cross entropy} (CE) loss is used for supervision:
\vspace{-1pt}
\begin{align}\label{equ:loss1}\small
\begin{split}
    \mathcal{L}^{mn}_{\text{basic}}\big((\bm{I}_m,\bm{I}_n), (\bm{l}_m,\bm{l}_n)\big)&\!=\!\mathcal{L}_{\text{CE}}(\bm{s}_m,\bm{l}_m)\!+\! \mathcal{L}_{\text{CE}}(\bm{s}_n,\bm{l}_n),\\
    &\!=\!\mathcal{L}_{\text{CE}}\big(\text{GAP}(\varphi(\bm{F}_{m})),\bm{l}_m\big)\!+\!\mathcal{L}_{\text{CE}} \big(\text{GAP}(\varphi(\bm{F}_{n})),\bm{l}_n\big).
    \end{split}
\vspace{-3pt}
\end{align}
So far the classifier is learned in a standard manner, \ie, only individual-image information is used for semantic learning. One can directly use the activation maps to supervise next-stage semantic segmentation learning, as done in \cite{dsrg2018,lee2019frame}. Differently, our classifier additionally utilizes a co-attention mechanism for further mining
cross-image semantics and eventually better localizing objects.

\noindent\textbf{Co-Attention for Cross-Image Common Semantics Mining.} Our co-attention attends to the two images, \ie, $\bm{I}_m$ and $\bm{I}_n$, simultaneously, and captures their correlations. %Note that here whether the two images are allowed to be same one.
We first compute the affinity matrix $\bm{P}$ between $\bm{F}_{m}$ and $\bm{F}_{n}$:
\vspace{-1pt}
\begin{align}\label{Eq:aff}\small
\begin{split}
        \bm{P}={\bm{F}_{m}^{\top}}\bm{W}_{\!\bm{P}}\bm{F}_{n} \in\!\mathbb{R}^{HW\!\times\!HW},
\end{split}
\vspace{-1pt}
\end{align}
where $\bm{F}_{m\!}\!\in\!\mathbb{R}^{C\!\times\!HW\!\!}$ and $\bm{F}_{n\!}\!\in\!\mathbb{R}^{C\!\times\!HW\!\!}$ are flattened into matrix formats, and $\bm{W}_{\!\bm{P}}\!\in\!\mathbb{R}^{C\!\times\!C\!}$ is a learnable matrix. The affinity matrix $\bm{P}$ stores similarity scores corresponding to all pairs of positions in $\bm{F}_{m}$ and $\bm{F}_{n}$, \ie, the $(i,j)^{th}$ element of $\bm{P}$ gives the similarity between $i^{th}$ location in $\bm{F}_{m}$ and $j^{th}$ location in $\bm{F}_{n}$.

Then $\bm{P}$ is normalized column-wise to derive attention maps across $\bm{F}_{m\!}$ for each position$_{\!}$ in$_{\!}$ $\bm{F}_{n}$, and$_{\!}$ row-wise$_{\!}$ to$_{\!}$ derive attention maps across $\bm{F}_{n\!}$ for each position in $\bm{F}_{m}$:
\vspace{-1pt}
\begin{align}\label{Eq:co-att}\small
\begin{split}
        \bm{A}_m\!=\!\text{softmax}(\bm{P}) \!\in\![0,1]^{HW\!\times\!HW\!},    ~~~\bm{A}_n\!=\!\text{softmax}(\bm{P}^{\!\top})\!\in\![0,1]^{HW\!\times\!HW\!},
\end{split}
\vspace{-1pt}
\end{align}
where softmax is performed column-wise.
In this way, $\bm{A}_n$ and $\bm{A}_m$ store the co-attention maps in their columns.
Next, we can compute attention summaries of $\bm{F}_{m}$ ($\bm{F}_{n}$) in light of each position of $\bm{F}_{n}$ ($\bm{F}_{m}$):
\vspace{-1pt}
\begin{align}\label{Eq:attsum}\small
\begin{split}
        \bm{F}^{m\cap n}_{m}=\bm{F}_{n}\bm{A}_n \!\in\!\mathbb{R}^{C\!\times\!H\!\times\!W},    ~~~~~\bm{F}^{m\cap n}_{n}=\bm{F}_{m}\bm{A}_m \!\in\!\mathbb{R}^{C\!\times\!H\!\times\!W},
\end{split}
\vspace{-1pt}
\end{align}
where $\bm{F}^{m\cap n\!}_{m}$ and $\bm{F}^{m\cap n}_{n}$ are reshaped into $\mathbb{R}^{C\!\times\!W\!\times\!H}$. %Each row of co-attention $\bm{A}_n$ stores the correspondence from all the entities of $\bm{F}_n$ to one position of $\bm{F}_{m}$. The computation of attention summaries $\bm{F}^{n\cap m}_{m}$ can be essentially viewed as interpreting each position of $\bm{F}_{m}$ by $\bm{F}_{n}$, and the common semantics can be preserved. Therefore,
Co-attentive feature $\bm{F}^{m\cap n}_m$, derived from $\bm{F}_{n}$, preserves the common semantics between $\bm{F}_{m\!}$ and $\bm{F}_{n\!}$ and locate the common objects in $\bm{F}_{m}$. Thus we can expect only the common semantics  $\bm{l}_{m\!}\cap\bm{l}_{n}$\footnote{The set operation `$\cap$' is slightly extended here to represent bitwise-and.} can be safely derived from $\bm{F}^{m\cap n}_{m}$, and  the same goes for $\bm{F}^{m\cap n}_{n}$. Such co-attention based common semantic classification can let the classifier understand the object patterns more completely and precisely.

To make things intuitive, consider the example in Fig.~\ref{fig:framework}, where $\bm{I}_m$ contains \textbf{Table} and \textbf{Person}, and $\bm{I}_n$ has \textbf{Cow} and \textbf{Person}.   As the co-attention is essentially the affinity computation between all the position pairs between $\bm{I}_m$ and $\bm{I}_n$, only the semantics of the common objects, \textbf{Person}, will be preserved in the co-attentive features, \ie, $\bm{F}^{m\cap n\!}_{m}$ and $\bm{F}^{m\cap n\!}_{n}$ (see Fig.~\ref{fig:framework}(b)). If we feed $\bm{F}^{m\cap n}_{m}$ and $\bm{F}^{m\cap n}_{n}$ into the class-aware fully convolutional layer $\varphi$, the generated class-aware activation maps, \ie, $\bm{S}^{m\cap n\!}_{m}\!=_{\!}\!\varphi(\bm{F}^{m\cap n\!}_{m})_{\!}\!\in_{\!}\!\mathbb{R}^{K\!\times\!H\!\times\!W\!\!}$ and $\bm{S}^{m\cap n\!}_{n}\!=_{\!}\!\varphi(\bm{F}^{m\cap n\!}_{n})_{\!}\!\in_{\!}\!\mathbb{R}^{K\!\times\!H\!\times\!W\!\!}$, are able to locate the common object \textbf{Person} in $\bm{I}_{m\!}$ and $\bm{I}_n$, respectively. After GAP, the predicted semantic classes (scores) $\bm{s}^{m\cap n}_{m}\!\in\!\mathbb{R}^{K\!}$ and $\bm{s}^{m\cap n}_{n}\!\in\!\mathbb{R}^{K\!}$ should be the common semantic labels $\bm{l}_{m\!}\cap\bm{l}_{n}$ of $\bm{I}_{m\!}$ and $\bm{I}_n$, \ie, \textbf{Person}.

Through co-attention computation, not only the human face, the most discriminative part of \textbf{Person}, but also other parts, such as legs and arms, are highlighted
in $\bm{F}^{m\cap n}_{m}$ and $\bm{F}^{m\cap n}_{n}$ (see Fig.~\ref{fig:framework}(b)). When we set the common class labels, \ie, \textbf{Person}, as the supervision signal, the classifier would realize that the semantics preserved in $\bm{F}^{m\cap n}_{m}$ and $\bm{F}^{m\cap n}_{n}$ are related and can be used to recognize \textbf{Person}.  Therefore, the co-attention, computed across two related images, \textit{explicitly} helps the classifier associate semantic labels and corresponding object regions and better understand the relations between different object parts. It essentially makes full use of the context across training data.

%For simplicity, we denote the computation of co-attention summaries $\bm{F}^{m\cap n}_m$ of $\bm{I}_m$ \wrt $\bm{I}_n$ as: $\bm{F}^{m\cap n}_m\!=\!\texttt{co-att}(\bm{I}_m, \bm{I}_n)$. Note that if $\bm{I}_m\!\neq\!\bm{I}_n$, $\texttt{co-att}(\bm{I}_m, \bm{I}_n)\!\neq\!\texttt{co-att}(\bm{I}_n, \bm{I}_m)$.

%When the paired training images $(\bm{I}_m,\bm{I}_n)$ are the \textit{same} one, \ie, $m\!=\!n$, the above conclusions are still tenable. In this case, the co-attention (Eq.~\ref{Eq:co-att}) computed between two same images can be viewed as  \textit{self-attention}~\cite{vaswani2017attention} or \textit{non-local operation}~\cite{wang2018non}.  As the correlations between all pixels (regions) of the same image are calculated by the co-attention,  intra-image foreground semantics and global self-context are explored. In addition, we adopt image processing based transformations during training. For $\bm{I}_m$,  a transformation operation $\tau$, randomly sampled from a set of pre-defined transformation operations $\mathcal{T}\!=\!\{\textsc{left-right flipping}, \times2~\textsc{scaling}, \times0.75~\textsc{scaling}\}$, is applied: $\tau(\bm{I}_m)$.  By applying image transformations, some originally discriminative regions may be distorted or presented in different scales, encouraging the classifier to seek more object patterns or richer context to make prediction,
%and further better identifying the entire extent of the objects.

Intuitively, for the co-attention based common semantic classification, the labels $\bm{l}_{m\!}\cap\bm{l}_n$ shared between $\bm{I}_m$ and $\bm{I}_n$ are used to supervise learning:
\begin{align}\label{equ:loss2}\small
\begin{split}
    \!\!\!\!\!\!\mathcal{L}^{mn}_{\text{co-att}\!}\big((\bm{I}_m,\bm{I}_{n\!}), (\bm{l}_{m}, \bm{l}_{n\!})\big)\!=&\mathcal{L}_{\text{CE}}(\bm{s}^{m\cap n\!}_{m},\bm{l}_{m\!}\!\cap\!\bm{l}_n) \!+\! \mathcal{L}_{\text{CE}}(\bm{s}^{m\cap n\!}_{n},\bm{l}_{m\!}\!\cap\!\bm{l}_n),\\
    \!=&\mathcal{L}_{\text{CE}}\big(\text{GAP}(\varphi(\bm{F}^{m\cap n}_{m})),\bm{l}_{m\!}\!\cap\!\bm{l}_{n\!}\big)+\\&\mathcal{L}_{\text{CE}} \big(\text{GAP}(\varphi(\bm{F}^{m\cap n}_{n})),\bm{l}_{m\!}\!\cap\!\bm{l}_{n\!}\big).\!\!\!\!\!\!
    \end{split}
\end{align}

\noindent\textbf{Contrastive Co-Attention for Cross-Image Exclusive Semantics Mining.} Aside from the co-attention described above that explores cross-image common semantics, we propose a
contrastive co-attention that mines semantic differences  between paired images. The co-attention and contrastive co-attention complementarily help the classifier better understand the concept of the objects.

As shown in Fig.~\ref{fig:framework}(a), for $\bm{I}_m$ and $\bm{I}_n$, we first derive \textit{class-agnostic co-attentions} from their  co-attentive features, \ie,  $\bm{F}^{m\cap n\!}_{m}$ and $\bm{F}^{m\cap n\!}_{n}$, respectively:
\begin{align}\label{Eq:ccm}\small
\begin{split}
        \!\!\bm{B}^{m\cap n}_{m}\!=\!\sigma(\bm{W}_{\bm{B}}\bm{F}^{m\cap n}_{m})\!\in~\!\![0,1]^{H\!\times\!W\!}, ~~\bm{B}^{m\cap n}_{n}\!=\!\sigma(\bm{W}_{\bm{B}}\bm{F}^{m\cap n}_{n})\!\in~\!\![0,1]^{H\!\times\!W\!},\!\!
\end{split}
\end{align}
where $\sigma(\cdot)$ is the \textit{sigmoid} activation function, and the parameter matrix $\bm{W}_{\bm{B}}\!\in\!\mathbb{R}^{1\!\times\! C}$ learns for common semantics collection and is implemented by a convolutional layer with$_{\!}$ $1\!\times\!1$$_{\!}$ kernel. $\bm{B}^{m\cap n}_{m\!}$ and $\bm{B}^{m\cap n}_{n\!}$ are$_{\!}$ class-agnostic$_{\!}$ and$_{\!}$ highlight$_{\!}$ all the common object regions$_{\!}$ in$_{\!}$ $\bm{I}_{m\!}$ and $\bm{I}_{n}$, respectively, based$_{\!}$ on$_{\!}$ which$_{\!}$ we$_{\!}$ derive$_{\!}$ contrastive$_{\!}$ co-attentions:$_{\!}$
\begin{align}\label{Eq:cca}\small
\begin{split}
        \bm{A}^{m\!\setminus\! n}_{m}=\bm{1}\!-\!\bm{B}^{m\cap n}_{m}\!\in~\!\![0,1]^{H\!\times\!W\!}, ~~~~\bm{A}^{n\!\setminus\! m}_{n}=\bm{1}\!-\!\bm{B}^{m\cap n}_{n}\!\in~\!\![0,1]^{H\!\times\!W\!}.
\end{split}
\end{align}
The contrastive co-attention $\bm{A}^{m\!\setminus\!n}_{m}$ of $\bm{I}_m$, as its superscript suggests, addresses those \textit{unshared}  object regions that are only of $\bm{I}_m$, but not of $\bm{I}_n$, and the same goes for $\bm{A}^{n\!\setminus\!m}_{n}$. Then we get \textit{contrastive co-attentive features}, \ie, unshared semantics in each images:
\begin{align}\label{Eq:cattsum}\small
\begin{split}
        \bm{F}^{m\!\setminus\! n}_{m}=\bm{F}_{m}\!\otimes\!\bm{A}^{m\!\setminus\! n}_{m}\!\in~\!\!\mathbb{R}^{C\!\times\! H\!\times\!W\!},    ~~~~~\bm{F}^{n\!\setminus\! m}_{n}=\bm{F}_{n}\!\otimes\!\bm{A}^{n\!\setminus\! m}_{n}\!\in~\!\!\mathbb{R}^{C\!\times\! H\!\times\!W\!}.
\end{split}
\end{align}
`$\otimes$' denotes element-wise multiplication, where the attention values are copied along the channel dimension.
Next, we can sequentially get class-aware activation maps, \ie, $\bm{S}^{m\!\setminus\! n\!}_{m}\!=_{\!}\!\varphi(\bm{F}^{m\!\setminus\! n}_{m})\!\in\!\mathbb{R}^{K\!\times\!H\!\times\!W}$ and $\bm{S}^{n\!\setminus\! m\!}_{n}\!=_{\!}\!\varphi(\bm{F}^{n\!\setminus\! m}_{n})\!\in\!\mathbb{R}^{K\!\times\!H\!\times\!W}$, and semantic scores, \ie, $\bm{s}^{m\!\setminus\! n\!}_{m}\!=\!\text{GAP}(\bm{S}^{m\!\setminus\! n}_{m})\!\in\!\mathbb{R}^{K\!}$ and $\bm{s}^{n\!\setminus\! m\!}_{n}\!=\!\text{GAP}(\bm{S}^{n\!\setminus\! m}_{n})\!\in\!\mathbb{R}^{K}$. For $\bm{s}^{m\!\setminus\! n}_{m}$ and $\bm{s}^{n\!\setminus\! m}_{n}$, they are expected to identify the categories of the unshared objects, \ie,  $\bm{l}_{m\!}\!\setminus\!\bm{l}_n$ and $\bm{l}_{n\!}\!\setminus\!\bm{l}_m$\footnote{The set operation `$\setminus$' is slightly extend here, \ie, $\bm{l}_{n\!}\!\setminus\!\bm{l}_m\!=\!\bm{l}_{n\!}-\bm{l}_{n\!}\!\cap\!\bm{l}_m$.}.

Compared with the co-attention that investigates common semantics as informative cues for boosting object patterns mining,
the contrastive co-attention addresses complementary knowledge from the semantic differences between paired images. Fig.~\ref{fig:framework}(b) gives an intuitive example.
After computing the contrastive co-attentions between $\bm{I}_m$ and $\bm{I}_n$  (Eq.~\ref{Eq:cca}), \textbf{Table} and \textbf{Cow}, which are unique in their original images, are highlighted. Based on the contrastive co-attentive features, \ie, $\bm{F}^{m\!\setminus\! n}_{m}$ and $\bm{F}^{n\!\setminus\! m}_{n}$, the classifier is required to accurately recognize \textbf{Table} and \textbf{Cow} classes, respectively. When the common objects are filtered out by the contrastive co-attentions, the classifier has a chance to focus more on the rest image regions and mine the unshared semantics more consciously. This also helps the classifier better discriminate the semantics of different objects, as the semantics of common objects and unshared ones are disentangled  by the contrastive co-attention. For example, if some parts of \textbf{Cow} are wrongly recognized as \textbf{Person}-related, the contrastive co-attention will discard these parts in $\bm{F}^{n\!\setminus\! m}_{n}$. However, the rest semantics in $\bm{F}^{n\!\setminus\! m}_{n}$ may be not sufficient enough for recognizing \textbf{Cow}. This will enforce the classifier to better discriminate different objects.

For the contrastive co-attention based unshared semantic classification, the supervision loss is designed as:
\begin{align}\label{equ:loss4}\small
\begin{split}
\!\!\!\!\mathcal{L}^{mn}_{\overline{\text{co-att}}}\big((\bm{I}_m,\bm{I}_{n\!}), (\bm{l}_{m}, \bm{l}_{n\!})\big)\!=&\mathcal{L}_{\text{CE}}(\bm{s}^{m\!\setminus\!n\!}_{m},\bm{l}_{m\!}\!\setminus\!\bm{l}_n) \!+\! \mathcal{L}_{\text{CE}}(\bm{s}^{n\!\setminus\!m\!}_{n},\bm{l}_{n\!}\!\setminus\!\bm{l}_m),\\
    \!=&\mathcal{L}_{\text{CE}}\big(\text{GAP}\big(\varphi(\bm{F}^{m\!\setminus\!n}_{m})\big),\bm{l}_{m\!}\!\setminus\!\bm{l}_n\big)\!+\\&\mathcal{L}_{\text{CE}} \big(\text{GAP}\big(\varphi(\bm{F}^{n\!\setminus\! m}_{n}\big),\bm{l}_{n\!}\!\setminus\!\bm{l}_m\big).\!\!\!\!
    \end{split}
\end{align}

\noindent\textbf{More In-Depth Discussion.} One can interpret our co-attention classifier from a view of \textit{auxiliary-task learning}~\cite{odena2017conditional,gidaris2018unsupervised}, which is investigated in self-supervised learning field to improve data efficiency and robustness, by exploring auxiliary tasks from inherent data structures. In our case, rather than the task of single-image semantic recognition which has been extensively studied in conventional WSSS methods, we explore two auxiliary
tasks, \ie, predicting the common and uncommon semantics from image pairs, for fully mining supervision signals from weak supervision. The classifier is driven to better understand the cross-image semantics by attending to (contrastive) co-attentive features, instead of only relying on intra-image information (see Fig.~\ref{fig:framework}(c)).  In addition, such strategy shares a spirit of \textit{image co-segmentation}. Since the image-level semantics of training set are given, the knowledge about some images share or unshare certain semantics should be used as a cue, or supervision signal, to better locate corresponding objects. Our co-attention based learning pipeline also provides an \textit{efficient data augmentation} strategy, due to the use of  paired samples, whose amount is near the square of the number of single training images.
%. The reasons are three-fold. First, pairs of training samples are used, growing as the square of the number of original training data. Second, %different but common-semantics shared images provide realistic and near-limitless object variants. For example, for the given training pairs $(\bm{I}_m,\bm{I}_n)$ in Fig. XX,  \textbf{Human} head in XX is occluded. When computing the co-attention summarizes from XX to XX,  semantics of human are missing. Thus the
%the classifier may need to pay more attention to other human patterns to successfully recognize \textbf{Human}. Meanwhile, through computing XX from XX, the correspondence between human head and other human parts is mined, leading to the comprehensive semantic understanding of \textbf{Human} class. Third, when the two training image samples are the same, we make different spatial transformations over them, pushing the classifier learning more complete knowledge of objects for transformation defense.

\vspace{-3pt}
\subsection{Co-Attention Classifier Guided WSSS Learning}\label{inference}
\vspace{-1pt}
\noindent\textbf{Training Co-Attention Classifier.} The overall training loss for  our co-attention  classifier ensembles the three terms defined in Eq.~\ref{equ:loss1},~\ref{equ:loss2}, and~\ref{equ:loss4}:
\vspace{-1pt}
\begin{align}\label{equ:loss5}\small
\begin{split}
\!\!\!\!\!\!\!\mathcal{L}\!=\!\!\sum\nolimits_{m,n}\!\mathcal{L}^{mn}_{\text{basic}}+\mathcal{L}^{mn}_{\text{co-att}}+
      \mathcal{L}^{mn}_{\overline{\text{co-att}}}.\!\!\!\!\!\!
    \end{split}
    \vspace{-3pt}
\end{align}
The coefficients of different loss terms are set as 1 in our all experiments. During training, to fully leverage the co-attention to mine the common semantics,  we sample two images $(\bm{I}_m, \bm{I}_n)$ with at least one common class, \ie, $\bm{l}_m\!\cap\!\bm{l}_n\!\neq\!\mathbf{0}$. %In addition, the contrastive co-attention based classification loss $\mathcal{L}^{mn}_{\overline{\text{co-att}}}$ is only active when paired-images are different, \ie, $m\!\neq\!n$.
%Furthermore, during training image pair sampling, the ratio between possibilities of sampling two same images and different ones is set as 3:1.

\noindent\textbf{Generating Object Localization Maps.} Once our image classifier is trained, we apply it over the training data $\mathcal{I}\!=\!\{(\bm{I}_n,\bm{l}_n)\}_n$ to produce corresponding object localization maps, which are essential for semantic segmentation network training. We explore two different strategies to generate localization maps.
\vspace*{-0pt}
\begin{itemize}[leftmargin=*]
	\setlength{\itemsep}{0pt}
	\setlength{\parsep}{-2pt}
	\setlength{\parskip}{-0pt}
	\setlength{\leftmargin}{-13pt}
	\vspace{-5pt}
    \item \textit{Single-round feed-forward prediction}, made over each training image individually.  For each training image $\bm{I}_n$, running the classifier and directly using its class-aware activation map  (\ie, $\bm{S}_{n\!}\!\in\!\mathbb{R}^{K\!\times\!H\!\times\!W}$) as the object localization map $\bm{L}_{n}$, as most previous network visualization based methods \cite{oaa2019,lee2019frame,shen2018bootstrapping} done.
	%\item \textit{Multi-round co-attentive prediction with self-augmentation}, which is made over each training image as well as its augmented ones. For $\bm{I}_n$, we use the three image transformations in $\mathcal{T}$, pre-defined in the training phase, to generate three transformed images $\{\bm{I}_{\tau(n)}\}_{\tau\in\mathcal{T}}$. Here $\bm{I}_{\tau(n)}$ is the abbreviation of $\tau(\bm{I}_n)$. Then we compute three co-attentions from $\{\bm{I}_{\tau(n)}\}_{\tau\in\mathcal{T}}$ to $\bm{I}_n$, and accumulate the corresponding class-aware activation maps as the final object localization map, \ie, $\bm{L}_{n\!}\!=_{\!}\!\frac{1}{|\mathcal{T}|}\!\sum_{\tau\in\mathcal{T}\!}\bm{S}^{m\cap \tau(n)}_{n}$. `$|\cdot|$' numerates the elements. Note that some prior art  \cite{hong2017weakly,lee2019frame,selvaraju2017grad} also combine activation maps from scaled and flipped images, but without considering self-correlations.
	\item \textit{Multi-round co-attentive prediction with extra reference information}, which is achieved by considering extra information from other related training images (see Fig.~\ref{fig:overview}(c)). Specifically, given a training image $\bm{I}_n$ and its associated label vector $\bm{l}_n$, we generate its localization map $\bm{L}_{n\!}$ in a \textit{class-wise} manner. For each semantic class $k\!\in\!\{1,\cdots,K\}$ labeled for $\bm{I}_n$, \ie, $l_{n,k}\!=\!1$ and $l_{n,k}$ is the $k^{th}$ element of $\bm{l}_n$, we sample a set of related images $\mathcal{R}\!=\!\{\bm{I}_r\}_r$ from $\mathcal{I}$, which are also annotated with label $k$, \ie, $l_{r,k}\!=\!1$. Then we compute the co-attentive feature $\bm{F}^{m\cap r}_{n\!}$ from each related image $\bm{I}_r\!\in\!\mathcal{R}$ to $\bm{I}_n$, and get the co-attention based class-aware activation map $\bm{S}^{m\cap r}_{n\!}$. Given all the class-aware activation maps $\{\bm{S}^{m\cap r}_{n\!}\}_r$ from $\mathcal{R}$, they are integrated to infer the localization map \textit{only} for class $k$, \ie, $L_{n,k\!}\!=\!\frac{1}{|\mathcal{R}|}\!\sum_{r\in\mathcal{R}\!}S^{m\cap r}_{n,k\!}\!$.
Here $L_{n,k\!}\!\in\!\mathbb{R}^{H\!\times\!W\!}$ and $S^{(\cdot)}_{n,k\!}\!\in\!\mathbb{R}^{H\!\times\!W\!}$ indicate the feature map at $k^{th\!}$ channel of $\bm{L}_{n\!}\!\in\!\mathbb{R}^{K\!\times\!H\!\times\!W\!}$ and $\bm{S}^{(\cdot)}_{n\!}\!\in\!\mathbb{R}^{K\!\times\!H\!\times\!W\!}$, respectively. `$|\cdot|$' numerates the elements.  After inferring the localization maps for all the annotated semantic classes of $\bm{I}_n$, we can get $\bm{L}_{n}$.
\vspace*{-2pt}
\end{itemize}

These two localization map generation strategies are studied in our experiments (\S\ref{sec:ablation}), and the last one is more favored, as it uses both intra- and inter-image semantics for object inference, and shares a similar data distribution of the training phase. One may notice that the contrastive co-attention is not used here. This is because contrastive co-attentive feature (Eq.~\ref{Eq:cattsum}) is from its original image, which is effective for boosting feature representation learning during classifier training, while contributes little for localization maps inference (with limited cross-image information). %In practice we do not observe significant performance change after using contrastive co-attention for object localization inference but the contrastive co-attention indeed boosts feature learning.
Related experiments can be found at \S\ref{sec:ablation}.

\noindent\textbf{Learning Semantic Segmentation Network.} After obtaining high-quality localization maps, we generate pseudo pixel-wise labels for all the training samples $\mathcal{I}$, which can be used to train arbitrary semantic segmentation network. %To make things comparable,
For pseudo groundtruth generation, we follow current popular pipeline~\cite{lee2019frame,lee2019ficklenet,oaa2019,hou2018self,dsrg2018,zeng2019joint}, that uses localization maps to extract class-specific object cues and adopts saliency maps \cite{HouPami19Dss,poolnet} to get background cues. For the semantic segmentation network, as in \cite{lee2019frame,lee2019ficklenet,oaa2019,hou2018self}, we choose DeepLab-LargeFOV~\cite{chen2017deeplab}.

\noindent\textbf{Learning with Extra Simple Single-Label Images.}  Some recent efforts \cite{pinheiro2015image,li2019attention}  are made towards exploring extra simple single-label images from other existing datasets \cite{russakovsky2015imagenet,griffin2007caltech} for further boosting WSSS. Though impressive, specific network designs are desired, due to the issue of domain gap between additionally used data and the target complex multi-label dataset, \ie, PASCAL VOC 2012~\cite{everingham2015pascal}. Interestingly, our co-attention based WSSS algorithm provides an alternate that addresses the challenge of domain gap naturally. Here we revisit the computation of co-attention in Eq.~\ref{Eq:aff}.
When $\bm{I}_m$ and $\bm{I}_n$ are from different domains, the parameter matrix $\bm{W}_{\!\bm{P}}$, in essence, learns to map them into a unified \textit{common semantic space}~\cite{an2020adversarial} and the co-attentive features can capture domain-shared semantics.  Therefore, for such setting, we learn three different parameter matrixes for $\bm{W}_{\!\bm{P}}$, for the cases where  $\bm{I}_m$ and $\bm{I}_n$ are from (1) the target semantic segmentation domain, (2) the one-label image domain, and (3) two different domains, respectively. Thus the domain adaption is efficiently achieved as a part of co-attention learning. We conduct related experiments in \S\ref{sec:ex-qp2}.

\noindent\textbf{Learning with Extra Web Images.}  Another trend of methods~\cite{jin2017webly,hong2017weakly,shen2018bootstrapping,wei2016stc}  address webly supervised semantic segmentation, \ie, leveraging web images as extra training samples. Though cheaper,  web data are typically noisy. To handle this,  previous arts propose diverse effective yet sophisticated solutions, such as multi-stage training~\cite{jin2017webly} and self-paced learning~\cite{wei2016stc}. Our co-attention based WSSS algorithm can be easily extended to this setting and solve data noise elegantly. As our co-attention classifier is trained with paired images, instead of previous methods  only relying on each image individually, our model provides a more robust training paradigm. In addition, during localization map inference, a set of extra related images are considered, which provides more comprehensive and accurate cues, and further improves the robustness. We$_{\!}$ experimentally$_{\!}$ demonstrate$_{\!}$ the$_{\!}$ effectiveness$_{\!}$ of$_{\!}$ our$_{\!}$ method$_{\!}$ in$_{\!}$ such$_{\!}$ a$_{\!}$ setting$_{\!}$ in$_{\!}$ \S\ref{sec:ex-qp3}.

\vspace{-6pt}
\subsection{Detailed Network Architecture}
\vspace{-2pt}
\noindent\textbf{Network Configuration.} In line with conventions~\cite{wei2018revisiting,zhang2018adversarial,oaa2019}, our image classifier is based on ImageNet~\cite{krizhevsky2012imagenet} pre-trained VGG-16 \cite{vgg}. For VGG-16 network, the last three fully-connected layers are replaced with three convolutional layers with 512 channels and kernel size $3\!\times\!3$, as done in \cite{oaa2019,zhang2018adversarial}.
% Then the {class-aware fully convolutional layer} $\varphi(\cdot)$ and {global average pooling} layer are further added to predict the semantics.
For the semantic segmentation network, for fair comparison with current top-leading methods \cite{oaa2019,ssdd2019,lee2019ficklenet,psa2018}, we adopt
the ResNet-101~\cite{he2016deep} version Deeplab-LargeFOV architecture.

\noindent\textbf{Training Phases of the Co-Attention Classifier and Semantic Segmentation Network. } % During training, training sample pairs are sampled and augmented as the way described in \S\ref{inference}.
Our co-attention classifier is fully end-to-end trained by minimizing the loss defined in Eq.~\ref{equ:loss5}. The training parameters are set as: initial learning rate (0.001) which is reduced by 0.1 after every 6 epochs, batch size (5), weight decay (0.0002), and momentum (0.9). Once the classifier is trained, we generate localization maps and pseudo segmentation masks over all the training samples   (see \S\ref{inference}). Then, with the masks,  the semantic segmentation network is trained in a standard way \cite{oaa2019} using the hyper-parameter setting in~\cite{chen2017deeplab}.

\noindent\textbf{Inference$_{\!}$ Phase$_{\!}$ of$_{\!}$ the$_{\!}$ Semantic$_{\!}$ Segmentation$_{\!}$ Network.} Given$_{\!}$ an$_{\!}$ \textit{unseen}$_{\!}$ test$_{\!}$ image, our$_{\!}$ segmentation$_{\!}$ network$_{\!}$ works$_{\!}$ in$_{\!}$ the$_{\!}$ \textit{standard}$_{\!}$ semantic$_{\!}$ segmentation$_{\!}$ pipeline$_{\!}$~\cite{chen2017deeplab}, \ie, directly$_{\!}$ generating$_{\!}$ segments$_{\!}$ without$_{\!}$ using any other images. Then CRF~\cite{krahenbuhl2011efficient} post-processing is performed to refine predicted masks.

Note that above settings are used in traditional WSSS datasets (\ie, \S\ref{sec:ex-qp1}, \S\ref{sec:ex-qp2}, \S\ref{sec:ex-qp3}). Due to the specific task setup in LID$_{20\!}$\!~\cite{lid2020}, corresponding training and testing settings will be detailed in \S\ref{sec:challenge}.

\vspace{-10pt}
\section{Experiment}
\label{sec:exp}
\vspace{-5pt}
\noindent\textbf{Overview.} Experiments
% \footnote{Additional details and results are available in supplementary material.}
 are first conducted over \textit{three} different WSSS settings: \textbf{(1)} The most standard paradigm \cite{oaa2019,ssdd2019,wei2017object,dsrg2018} that only allows image-level supervision from PASCAL VOC 2012~\cite{everingham2015pascal} (see \S\ref{sec:ex-qp1}). \textbf{(2)} Following \cite{li2019attention,pinheiro2015image}, additional single-label images can be used, yet bringing the challenge of domain gap (see \S\ref{sec:ex-qp2}). \textbf{(3)} Webly supervised semantic segmentation paradigm \cite{lee2019frame,shen2018bootstrapping,jin2017webly}, where extra web data can be accessed (see \S\ref{sec:ex-qp3}). Then, in \S\ref{sec:challenge}, we show the results in WSSS track of LID$_{20\!}$, where our method achieves the champion. Finally, in \S\ref{sec:ablation}, ablation studies are made to assess the effectiveness of essential parts of our algorithm.

% \vspace{-6pt}

\noindent\textbf{Evaluation Metric.} In our experiments, the standard intersection over union (IoU) criterion is reported on the val and test sets of PASCAL VOC 2012~\cite{everingham2015pascal}. The scores on test set are obtained from official PASCAL VOC evaluation server.

\vspace{-8pt}
\subsection{Experiment 1: Learn WSSS only from PASCAL VOC ~\cite{everingham2015pascal} Data} \label{sec:ex-qp1}
\vspace{-3pt}
\noindent\textbf{Experimental Setup:} We first conduct experiment following the most standard setting that learns WSSS with only image-level labels~\cite{oaa2019,ssdd2019,wei2017object,dsrg2018}, \ie, only image-level supervision from PASCAL VOC 2012~\cite{everingham2015pascal} is accessible. PASCAL VOC 2012 contains a total of 20 object categories. As in~\cite{chen2017deeplab,wei2017object}, augmented training data from \cite{hariharan2011semantic} are also used. Finally, our model is trained on totally 10,582 samples with only image-level annotations.  Evaluations are conducted on the val and test sets, which have 1,449 and 1,456 images, respectively.

%\noindent\textbf{Implementation Details:} The fully segmentation network is trained till 15000 iterations, following \cite{chen2017deeplab}.
\begin{table}[t]
\captionsetup{font=small}
\caption{\small{Experimental results for WSSS under three different settings. \textbf{(a)} Standard setting where only PASCAL VOC 2012 images are used (\S\ref{sec:ex-qp1}). \textbf{(b)} Additional single-label images are used (\S\ref{sec:ex-qp2}). \textbf{(c)} Additional web-crawled images are used (\S\ref{sec:ex-qp3}). *:~VGG backbone. \dag: ResNet backbone.}}
\begin{subtable}{0.5\linewidth}
\centering
% \begin{tabular}{|c|c|c|c|}
% \hline
% text & text & text & text \\ \hline
% text & text & text & text \\ \hline
% \end{tabular}
% \caption{SUBTBL1}
\resizebox{1\textwidth}{!}{
	\setlength\tabcolsep{8pt}
	\renewcommand\arraystretch{1.07}
        \begin{tabular}{l||c|c|c}
\hline\thickhline
\rowcolor{mygray}
~~~~~~Methods                & Publication&Val                       & Test                      \\ \hline\hline
\multicolumn{4}{c}{\textbf{Using PASCAL  VOC data only}} \\\hline\hline
% CCNN~\cite{}  $_\text{ICCV15}$     & 35.3                      & 35.6                      \\ \hline
% EM-Adapt~\cite{papandreou2015weakly}  &$\text{ICCV15}$ & 38.2                      & 39.6                      \\
*DCSM~\cite{wataru2016distinct} &$\text{ECCV16}$      & 44.1                      & 45.1                      \\ %\hline
*SEC~\cite{sec2016} &$\text{ECCV16}$    & 50.7                      & 51.7                      \\ %\hline
*AFF~\cite{qi2016augmented} &$\text{ECCV16}$       & 54.3                      & 55.5                      \\% \hline
\dag DCSP~\cite{chaudhry2017discovering} &$\text{BMVC17}$      & 60.8                      & 61.9                      \\ %\hline
*CBTS~\cite{roy2017combining} &$\text{CVPR17}$      & 52.8                      & 53.7                      \\ %\hline
*AE-PSL~\cite{wei2017object} &$\text{CVPR17}$    & 55.0                        & 55.7                      \\ %\hline
*Oh \etal~\cite{hong2017weakly} &$\text{CVPR17}$ & 55.7                      & 56.7                      \\% \hline
**TPL~\cite{kim2017two} &$\text{ICCV17}$       & 53.1                      & 53.8                      \\ %\hline
*MEFF~\cite{meff_2018} &$\text{CVPR18}$     & -                         & 55.6                      \\ %\hline
*GAIN~\cite{li2018tell} &$\text{CVPR18}$      & 55.3                      & 56.8                      \\ %\hline
*MDC~\cite{wei2018revisiting} &$\text{CVPR18}$       & 60.4                      & 60.8                      \\ %\hline
\dag MCOF~\cite{mcof_2018} &$\text{CVPR18}$      & 60.3                      & 61.2                      \\ %\hline
\dag DSRG~\cite{dsrg2018} &$\text{CVPR18}$      & 61.4                      & 63.2                      \\ %\hline
\dag PSA~\cite{psa2018} &$\text{CVPR18}$       & 61.7                      & 63.7                      \\ %\hline
\dag SeeNet~\cite{hou2018self} &$\text{NIPS18}$    & 63.1                      & 62.8                      \\ %\hline
\dag IRN~\cite{irn_2019_CVPR} &$\text{CVPR19}$       & 63.5                      & 64.8                      \\ %\hline
\dag FickleNet~\cite{lee2019ficklenet} &$\text{CVPR19}$ & 64.9                      & 65.3                      \\ %\hline
\dag SSDD~\cite{ssdd2019} &$\text{ICCV19}$      & 64.9                      & 65.5                      \\ %\hline
\dag OAA+~\cite{oaa2019} &$\text{ICCV19}$    & 65.2                      & 66.4                      \\ %\hline
\hline\hline
*\textbf{Ours} (VGG)          &-         &63.5 &63.6 \\
\dag\textbf{Ours} (ResNet)          &-         & \multicolumn{1}{l|}{\textbf{66.2}} & \multicolumn{1}{l}{\textbf{66.9}} \\\hline
\end{tabular}}
\vspace{-0.5cm}
\captionsetup{font={small}}
\caption{}
\label{results:wo_data}
\end{subtable}%
\begin{subtable}{0.5\linewidth}
\centering
\resizebox{1\textwidth}{!}{
	\setlength\tabcolsep{8pt}
        \begin{tabular}{l||c|c|c}
\hline\thickhline
\rowcolor{mygray}
Methods             &Publication & Val  & Test \\ \hline\hline
\multicolumn{4}{c}{\textbf{Using extra simple single-label images}} \\\hline\hline
*MCNN~\cite{tokmakov2016weakly} &$\text{ICCV15}$        & - & 36.9 \\
*MIL-ILP~\cite{pinheiro2015image} &$\text{CVPR15}$ & 32.6   & - \\
*MIL-sppxl~\cite{pinheiro2015image} &$\text{CVPR15}$ & 36.6   & 35.8 \\
*MIL-bb~\cite{pinheiro2015image} &$\text{CVPR15}$ & 37.8   & 37.0 \\
*MIL-seg~\cite{pinheiro2015image} &$\text{CVPR15}$ & 42.0   & 40.6 \\
*AttnBN~\cite{li2019attention} &$\text{ICCV19}$ & 62.1 & 63.0   \\ \hline\hline
*\textbf{Ours} (VGG)          &-         &64.6 &64.6 \\
\dag\textbf{Ours} (ResNet)             &-   &  \textbf{67.1}    &   \textbf{67.2}   \\ \hline
\end{tabular}}
\vspace{-0.5cm}
\captionsetup{font={small}}
\caption{}
\label{results:w_single}

% \vspace{-0.101cm}
%\vspace{0.05cm}
\resizebox{1\textwidth}{!}{
\setlength\tabcolsep{7.2pt}
\begin{tabular}{l||c|c|c}
\hline\thickhline
\rowcolor{mygray}
Methods                  & Publication &Val  & Test \\ \hline \hline
\multicolumn{4}{c}{\textbf{Using extra noisy web images/videos}} \\\hline\hline
*MCNN~\cite{tokmakov2016weakly} &$\text{ICCV15}$        & 38.1 & 39.8 \\
\dag Shen \etal~\cite{shen2017weakly} & $\text{BMVC17}$  & 56.4   & 56.9 \\
*STC~\cite{wei2016stc}  &$\text{PAMI17}$         & 49.8 & 51.2 \\
*Hong \etal~\cite{hong2017weakly} & $\text{CVPR17}$  & 58.1 & 58.7 \\
*WebS-i1~\cite{jin2017webly} & $\text{CVPR17}$      & 51.6 & - \\
*WebS-i2~\cite{jin2017webly} & $\text{CVPR17}$      & 53.4 & 55.3 \\
\dag Shen \etal~\cite{shen2018bootstrapping} & $\text{CVPR18}$  & 63.0   & 63.9 \\
% Lee \etal~\cite{lee2019frame} & $\text{ICCV19}$   & 66.5 & 67.4 \\
\hline\hline
*\textbf{Ours} (VGG)          &-         &65.0 &64.7 \\
\dag\textbf{Ours}  (ResNet)            &-         &  \textbf{67.7}    &   \textbf{67.5}   \\ \hline
\end{tabular}}
\vspace{-0.5cm}
\captionsetup{font={small}}
\caption{}
\label{results:w_web}
\end{subtable}
\vspace{-24pt}
\end{table}

\begin{figure}[t]
\includegraphics[width=1\linewidth, clip=true, trim=0cm 15.5cm 2.6cm 0cm]{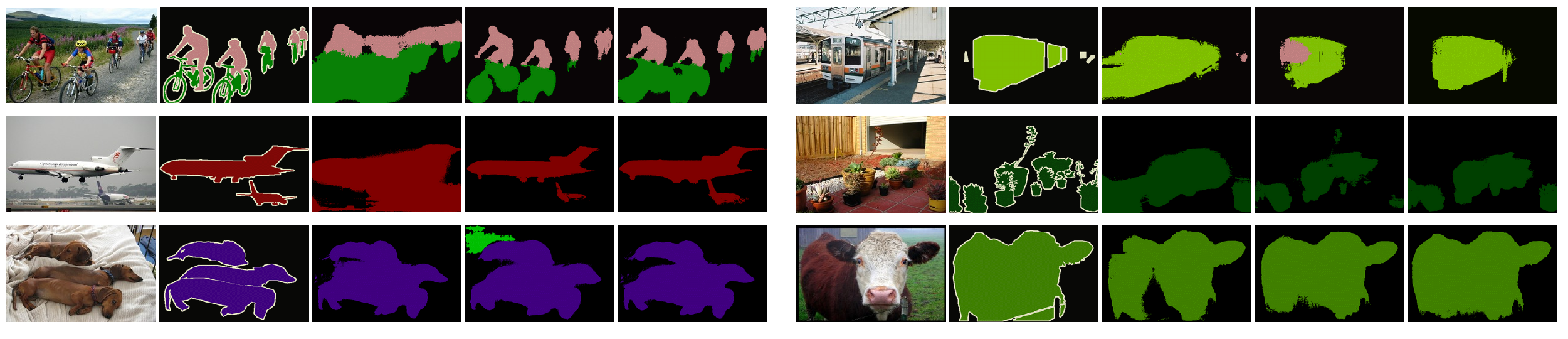}\\  \vspace{-0.30 cm}%3.5\\
			%\hspace*{0.25\linewidth} (a) \hspace*{0.\linewidth}(b) \\
			 \vspace*{-0.3cm}
\captionsetup{font=small}
			\caption{\small{Visual comparison results on PASCAL VOC12 val set. From \textit{left} to \textit{right}: input image, ground truth, results for PSA \cite{psa2018}, OAA+ \cite{oaa2019}, and our method.}}
			\label{fig:qual}
			 \vspace*{-0.5cm}
\end{figure}

\noindent\textbf{Experimental Results:} Table \ref{results:wo_data} compares our approach and current top-leading WSSS methods (highest mIoU is used for comparison) with image-level supervision, on both PASCAL VOC12 val and test sets. Additionally, we show some segmentation results in Fig. \ref{fig:qual}. We can observe that our method achieves mIoU scores of 66.2 and 66.9 on val and test sets respectively, outperforming all the competitors. The performance of our method is 87\% of the DeepLab-LargeFOV \cite{chen2017deeplab} trained with fully annotated data, which achieved an mIoU of 76.3 on val set. When compared to OAA+ \cite{oaa2019}, current best-performing method, our approach obtains the improvement of 1.0\% on val set. This well demonstrates that the localization maps produced by our co-attention classifier effectively detect more complete semantic regions towards the whole target objects. Note that our network is elegantly trained end-to-end in a single phase. In contrast, many other recent approaches including OAA+~\cite{oaa2019} and SSDD~\cite{ssdd2019}, use extra networks \cite{oaa2019,ssdd2019,psa2018} to learn auxiliary information (\eg, integral attention~\cite{oaa2019}, pixel-wise semantic affinity~\cite{ssdd2019}, \etc), or adopt multi-step training \cite{irn_2019_CVPR,wei2017object,wei2018revisiting}.

 \vspace{-5pt}
\subsection{\!Experiment$_{\!}$ 2: Learn$_{\!}$ WSSS$_{\!}$ with$_{\!}$ Extra$_{\!}$ Simple$_{\!}$ Single-Label$_{\!}$ Data$_{\!}$} \label{sec:ex-qp2}
 \vspace{-3pt}
\noindent\textbf{Experimental Setup:} Following \cite{pinheiro2015image,li2019attention}, we train our co-attention classifier and segmentation network with PASCAL images and extra single-label images. The extra single-label images are borrowed from the subsets of Caltech-256 \cite{griffin2007caltech} and ImageNet CLS-LOC \cite{russakovsky2015imagenet}, and whose annotations are within 20 VOC object categories. There are a total of 20,057 extra single-label images.

%\noindent\textbf{Implementation Details:} Since more images are used, we train the segmentation network by 20000 iterations.

\noindent\textbf{Experimental Results:} The comparisons are shown in Table \ref{results:w_single}. Our method significantly improves the most recent method (\ie, AttnBN~\cite{li2019attention}) in this setting by 5.0\% and 4.2\% in val and test sets, respectively. With the fact that objects of the same category but from different domains share similar visual patterns~\cite{li2019attention}, our co-attention provides an end-to-end strategy that  efficiently captures the common, cross-domain semantics, and learns domain adaption naturally. Even AttnBN is specifically designed for addressing such setting by knowledge transfer, our method still suppresses it by a large margin. Compared with the setting in \S\ref{sec:ex-qp1} where only PASCAL images are used for training, our method obtains improvements on both val and test sets, verifying that it successfully mines knowledge from extra simple single-label data and copes with domain gap well.

% \begin{table}[!t]
% \caption{Comparison of the performance of WSSS with additional single-label images on PASCAL VOC12 validation and test sets.}
% \label{results:w_single}
% \centering\small
% 	\resizebox{0.45\textwidth}{!}{
% 	\setlength\tabcolsep{6pt}
% 	\renewcommand\arraystretch{1.0}
% \begin{tabular}{l||c|c|c}
% \hline\thickhline
% \rowcolor{mygray}
% Methods             &Publication & Val  & Test \\ \hline\hline
% MIL-seg~\cite{pinheiro2015image} &$\text{CVPR15}$ & 42.0   & 40.6 \\
% AttnBN~\cite{li2019attention} &$\text{ICCV19}$ & 62.1 & 63.0   \\ \hline\hline
% \textbf{Ours}              &-   &  \textbf{66.9}    &   \textbf{67.2}   \\ \hline
% \end{tabular}}
% \end{table}
 % \vspace{-0pt}
 % \vspace{5pt}

\subsection{Experiment 3: Learn WSSS with Extra Web-Sourced Data} \label{sec:ex-qp3}
 \vspace{-3pt}
\noindent\textbf{Experimental Setup:} We also conduct experiments using both PASCAL VOC images and webly craweled images as training data. We use the web data provided by \cite{shen2018bootstrapping}, which are retrieved from Bing based on class names. %, and roughly cleaned using Initial-SEC \cite{shen2018bootstrapping} and setting thresholds for each class.
% After roughly cleaning process,
The final dataset contains 76,683 images across 20 PASCAL VOC classes.

%\noindent\textbf{Implementation Details:} Here, we also train the segmentation network by 20000 iterations.

\begin{table}[t]
	\centering\small
\captionsetup{font=small}
	\caption{\small{Ablation study for different object localization map generate strategies, reported on PASCAL VOC12 val set. See \S\ref{sec:ablation} for details.}}
	\label{table:inference_mode}
	%\vspace*{-8pt}
%\begin{threeparttable}
	\resizebox{0.85\textwidth}{!}{
	\setlength\tabcolsep{8pt}
	\renewcommand\arraystretch{1.0}
	\begin{tabular}{c|c|c||c}
		\hline\thickhline
				\rowcolor{mygray}
Method&Inference Mode&Input Image(s) &Val \\  \hline \hline
\multirow{1}{*}{Basic Classifier}&Single-round feed-forward &Test image \textit{only}& 61.7\\
%\cline{2-4}
%&Multi-round feed-forward &Self-augmented images& 59.8\\
\hline\hline
\multirow{3}{*}{Our Variant}&Single-round  feed-forward&Test image \textit{only}& 64.7\\
%\cline{2-4}
%&Multi-round co-attention &Self-augmented images&65.8\\
\cline{2-4}
& Multi-round co-attention &Test image&\multirow{2}{*}{66.2}\\
\specialrule{0em}{-0.5pt}{-1.5pt}
& and contrastive co-attention&and other related images&\\
 \hline\hline
\multirow{2}{*}{\textbf{Full Model}}&\multirow{2}{*}{Multi-round co-attention}&Test image&\multirow{2}{*}{\textbf{66.2}}\\
\specialrule{0em}{-0.5pt}{-1.5pt}
&&and other related images&\\
\hline
\end{tabular}
}
%\end{threeparttable}
\vspace{-12pt}
\end{table}

\noindent\textbf{Experimental Results:} Table \ref{results:w_web} shows the performance comparisons between our method and the previous webly supervised segmentation methods. It shows that our method outperforms all other approaches and sets new state-of-the-arts with mIoU score of 67.7 and 67.5 on PASCAL VOC 2012 val and test sets, respectively. Among the compared methods, Hong \etal \cite{hong2017weakly} utilize richer information of the temporal dynamics provided by additional large-scale videos. In contrast, although only using static image data, our method still outperforms it on the val and test sets by 9.6\% and 8.8\%, respectively. Compared with Shen \etal \cite{shen2018bootstrapping} which uses the same web data as ours, our method substantially improves it by a clear margin of 3.6\% on the test set.

\subsection{Experiment 4: Performance on WSSS Track of LID$_{20\!}$ Challenge} \label{sec:challenge}
\noindent\textbf{Experimental Setup:} The challenge dataset \cite{lid2020} is built upon ImageNet \cite{russakovsky2015imagenet}. It contains 349,319 images with image-level labels from 200 classes. Evaluations are conducted on the val and test sets, which have 4,690 and 10,000 images, respectively. In this challenge, our co-attention image classifier is built upon ResNet-38\cite{wu2019wider}, as the dataset has 200 classes and a stronger backbone can better learn subtle semantics between classes. The training parameters are set as: initial learning rate (0.005) and the poly policy based training schedule: $lr\!=\!lr_{init}\!\times\!(1\!-\!\frac{iter}{max\_iter})^\gamma$ with $\gamma$(0.9), batch size (8), weight decay (0.0005), and max epoch (15). During training, the equivariant attention~\cite{seam} is also adopted. Once our image classifier is trained, we run the classifier and directly use its class-aware activation map (\ie, $\bm{S}_{n}$) as the object localization map $\bm{L}_{n}$. Then we generate pseudo pixel-wise labels for all the training samples $\mathcal{I}$. Since only image tags can be used, we follow\!~\cite{psa2018}: localization maps are first used to train an AffinityNet model, which is then used to generate pseudo ground truth masks and background threshold is set as 0.2. For better segmentation results, we choose ResNet-101 based DeepLab-V3. The parameters are set as below: initial learning rate (0.007) with poly schedule, batch size (48), max epoch (100), and weight decay (0.0001). The segmentation model is trained on 4 Tesla V100 GPUs. During testing, results from multiple scales are averaged, with CRF refinement.

\noindent\textbf{Experimental Results:} The final results with the standard mean intersection over union (mIoU) criterion for WSSS track of both LID$_{19\!}$ and LID$_{20\!}$ challenges are shown in Table\!~\ref{table:lid_results}. Both LID$_{19\!}$ and LID$_{20\!}$ challenge use the same data. In LID$_{19}$, competitors can use extra saliency annotations to learn saliency models and refine pseudo ground truths. However, in LID$_{20}$, only image tags can be accessed. For methods shown in the table, top performing methods are included. As can be seen from Table \ref{table:lid_results}, our approach not only outperforms the champion team in LID$_{19}$, which can use deep learning based saliency models, but also achieves the best performance in LID$_{20\!}$ and sets a new state-of-the-art (\ie, mIoU of 46.2 and 45.1 in val and test sets, respectively). %Some qualitative results on validation set are shown in Fig. \ref{fig:qual-lid}.

\begin{table}[t]
  \centering
%\vspace{-7pt}
\captionsetup{font=small}
  \caption{\small {Results on \textit{val} and \textit{test} sets of both LID$_{19\!}$ and LID$_{20\!}$ WSSS track.}}
  \label{table:lid_results}
  %\vspace*{-8pt}
%\begin{threeparttable}
  \resizebox{0.9\textwidth}{!}{
\setlength\tabcolsep{10pt}
  \begin{tabular}{c|c|c|c|c}
    \hline\thickhline
        \rowcolor{mygray}
Year&Team &Extra Saliency Annotation&Val&Test\\
\hline \hline
\multirow{3}{*}{LID$_{19\!}$}& T.T (T.T) & \cmark& - &8.1\\\cline{2-5}
& LEAP\_DEXIN& \cmark&20.7 & 19.6\\\cline{2-5}
& MVN & \cmark& 41.0& 40.0\\\hline\hline
\multirow{5}{*}{LID$_{20\!}$}&play-njupt  &\xmark &22.1 & 31.9\\\cline{2-5}
&IOnlyHaveSevenDays &\xmark &39.0& 36.2\\\cline{2-5}
&UCU \& SoftServe  &\xmark&39.7 & 37.3\\\cline{2-5}
&VL-task1 & \xmark&40.1& 37.7\\ \cline{2-5}
&CVL (\textbf{ours})& \xmark& \textbf{46.2}& \textbf{45.1}\\
\hline
\end{tabular}
}
%\end{threeparttable}
\vspace{-14pt}
\end{table}

\subsection{Ablation Studies} \label{sec:ablation}
 \vspace{-1pt}
\noindent\textbf{Inference Strategies.} Table \ref{table:inference_mode} shows mIoU scores on PASCAL VOC 2012 val set \wrt~different inference modes (see \S\ref{inference}). When using the traditional inference mode ``single-round feed-forward", our method substantially suppresses basic classifier, by improving mIoU score from 61.7 to 64.7. This evidences that co-attention mechanism (trained in an end-to-end manner) in our classifier improves the underlying feature representations and more object regions are identified by the network. We can observe that by using more images to generate localization maps, our method obtains consistent improvement from ``Test image \textit{only}'' (64.7), to ``Test images and other related images'' (66.2). This is because more semantic context are exploited during localization map inference. In addition, using contrastive co-attention for localization map inference doesn't boost performance (66.2). This is because the contrastive co-attentive features for one image are derived from the image itself. In contrast, co-attentive features are from the other related image, thus can be effective in the inference stage. %Hence, adding contrastive co-attentive features doesn't bring more information.

\noindent\textbf{(Contrastive) Co-Attention.} As seen in Table \ref{table:contrastive_co_atten}, by only using co-attention (Eq.~\ref{equ:loss2}), we already largely suppress the basic classifier (Eq.~\ref{equ:loss1}) by 3.8\%. When adding additional contrastive co-attention (Eq.~\ref{equ:loss4}), we obtain mIoU improvement of 0.7\%. Above analysis verify our two co-attentions indeed boost performance.
%\noindent\textbf{Training Transformations.} We adopt different image transformations during training to mine the global context within the images (see \S\ref{co-cls}) and show the performance comparisons in Table \ref{table:transformations_training}. From second row to the last row in Table \ref{table:transformations_training}, more complex transformations are used and more obejct patterns are identified, leading to consistent performance gain.

\begin{table}[t]
	\centering\small
\captionsetup{font=small}
	\caption{\small{Ablation study for our co-attention and contrastive co-attention mechanisms for training, reported on PASCAL VOC12 val set. See \S\ref{sec:ablation} for details.}}
	\label{table:contrastive_co_atten}
	%\vspace*{-8pt}
%\begin{threeparttable}
	\resizebox{0.95\textwidth}{!}{
	\setlength\tabcolsep{6pt}
	\renewcommand\arraystretch{1}
	\begin{tabular}{c|c|c||c}
		\hline\thickhline
				\rowcolor{mygray}
Method&(Contrastive) Co-Attention&Training Loss&Val\\  \hline \hline
Basic Classifier&-&$\mathcal{L}_{\text{basic}}$ (Eq.~\ref{equ:loss1})& 61.7\\ \hline\hline
\multirow{1}{*}{Our Variant}& co-attention \textit{only}&$\mathcal{L}_{\text{basic}}$ (Eq.~\ref{equ:loss1})+$\mathcal{L}_{\text{co-att}}$ (Eq.~\ref{equ:loss2})& 65.5\\\hline\hline
\multirow{2}{*}{\textbf{Full Model}}&co-attention&$\mathcal{L}_{\text{basic}}$ (Eq.~\ref{equ:loss1})+$\mathcal{L}_{\text{co-att}}$ (Eq.~\ref{equ:loss2})+$\mathcal{L}_{\overline{\text{co-att}}}$ (Eq.~\ref{equ:loss4})&\multirow{2}{*}{\textbf{66.2}}\\
&+contrastive co-attention&= $\mathcal{L}$ (Eq.~\ref{equ:loss5})& \\
\hline
\end{tabular}
}
%\end{threeparttable}
\vspace{-10pt}
\end{table}
\begin{wraptable}{r}{6cm}
	\centering	%
\captionsetup{font=small}
\vspace*{-20pt}
	\caption{Ablation study for using different numbers of related images during object localization map generation, reported on PASCAL VOC12 val set (see \S\ref{sec:ablation}).}
	\label{table:num_extraed_images}
\vspace*{-6pt}
	\resizebox{0.48\textwidth}{!}{
	\setlength\tabcolsep{4pt}
	\renewcommand\arraystretch{1.0}
	\begin{tabular}{l|c||c}
		\hline\thickhline
				\rowcolor{mygray}
				% &Extra Related &\\
				\rowcolor{mygray}
% \multirow{-2}{*}{Method}&Images (\#)~~~~&\multirow{-2}{*}{Val}\\  \hline \hline
{Method}& Extra Related Images (\#)&~~{Val}~~~~\\  \hline \hline
\multirow{5}{*}{Our Variant}& 0 &64.7\\
\cline{2-3}
& 1 &65.9\\
\cline{2-3}
& 2 &66.0\\
\cline{2-3}
&4&66.1\\
\cline{2-3}
&5&66.0\\
\hline\hline
\textbf{Full Model}&3&\textbf{66.2}\\
\hline
\end{tabular}
}
%\end{threeparttable}
\vspace{-12pt}
\end{wraptable}

% \vspace*{-9pt}
\noindent\textbf{Number of Related Images for Localization Map Inference.} For localization map generation, we use 3 extra related images (\S\ref{inference}). Here, we study how the number of reference images affect the performance. From Table \ref{table:num_extraed_images}, it is easily observed that when increasing the number of related images from 0 to 3, the performance gets boosted consistently. However, when further using more images, the performance degrades. This can be attributed to the trade-off between useful semantic information and noise brought by related images. From 0 to 3 reference images, more semantic information is used and more integral regions for objects are mined. When further using more related images, useful information reaches its bottleneck and noise, caused by imperfect localization of the classifier, takes over, decreasing performance.

% \begin{table}[t]
% 	\centering\small
% %\captionsetup{font=small}
% 		\caption{Ablation study for adopting image transformations during training, reported on PASCAL VOC12 val set. See \S\ref{sec:ablation} for details.}
% 	\label{table:transformations_training}
% 	%\vspace*{-8pt}
% %\begin{threeparttable}
% 	\resizebox{0.6\textwidth}{!}{
% 	\setlength\tabcolsep{4pt}
% 	\renewcommand\arraystretch{1.0}
% 	\begin{tabular}{l|c||c}
% 		\hline\thickhline
% 				\rowcolor{mygray}
% Method&Training Strategy&Val \\  \hline \hline
% \multirow{3}{*}{Our Variant}&\textit{w/o.} image transformation &\\
% \cline{2-3}
% &\textit{w.} flip \textit{only}&\\
% \cline{2-3}
% &\textit{w.} scale \textit{only}&\\ \hline\hline
% \textbf{Full Model}&flip + scale&\\
% \hline
% \end{tabular}
% }
% %\end{threeparttable}
% \vspace{-10pt}
% \end{table}
% \vspace{-5pt}

% \noindent\textbf{Visual Comparisons.}
\vspace{-7pt}
\section{Conclusion}
\vspace{-5pt}
% In this work, we propose a co-attention classification network to discover more integral object regions, by considering cross-image semantic relations between training images. With this regard, co-attention mechanism is exploited to mine the common semantic regions within paired samples, so that complete object patterns are learned. Additionally, we utilize contrastive co-attention to focus on the exclusive and unshared ones, for capturing complimentary supervision cues. Extensive experiments are conducted under three settings. For all cases, new state-of-the-art performance is achieved, demonstrating the effectiveness of our method. Further, by exploiting additional single-label images and web-crawled images, our approach is proven to generalize well under domain gap and data noise.
This work proposes a co-attention classification network to discover integral object regions by addressing cross-image semantics. With this regard, a co-attention is exploited to mine the common semantics within paired samples, while a contrastive co-attention is utilized to focus on the exclusive and unshared ones for capturing complimentary supervision cues. Additionally, by leveraging extra context from other related images, the co-attention boosts localization map inference. Further, by exploiting additional single-label images and web images, our approach is proven to generalize well under domain gap and data noise. Experiments over three WSSS settings consistently show promising results. Our method also ranked 1$^{st\!}$ place in the weakly-supervised semantic segmentation track of LID$_{20\!}$ challenge.

{\small
\bibliographystyle{splncs04}
\bibliography{egbib}
}
\end{document}